# Improvement of Collision Avoidance in Cut-In Maneuvers Using Time-to-Collision Metrics

**Jamal Raiyn**

Computer Science department, Al Qasemi Academic College, Baka Al Gharbiya, Israel

raiyn@qsm.ac.il

**Abstract:** This paper proposes a new strategy for collision avoidance system leveraging Time-to-Collision (TTC) metrics for handling cut-in scenarios, which are particularly challenging for autonomous vehicles (AVs). By integrating a deep learning with TTC calculations, the system predicts potential collisions and determines appropriate evasive actions compared to traditional TTC -based approaches. The methodology is validated through extensive simulations, demonstrating a significant improvement in collision avoidance performance compared to traditional TTC-based approaches. By combining deep learning models with TTC calculations, the system predicts potential collisions and determines appropriate evasive actions. The methodology is validated through simulations, demonstrating significant improvements in collision avoidance performance. The use of the Gaussian model to contributes to Time-to-Collision (TTC) analysis by providing a probabilistic framework to quantify collision risk under uncertainty. It calculates the likelihood that TTC will fall below a critical threshold (TTC_crit), indicating a potential collision. By modeling input variations—such as sensor inaccuracies, fluctuating vehicle velocity, and unpredictable driving behavior—as a Gaussian distribution, the system can handle real-world uncertainties more effectively. This enables continuous, real-time risk prediction, allowing for dynamic and adaptive collision avoidance decisions. The Gaussian approach enhances the robustness of TTC-based systems by improving their ability to predict and prevent collisions in uncertain driving conditions.

## 1.    Introduction

The rapid growth in autonomous vehicle (AV) technology has introduced the need for advanced safety systems capable of preventing accidents in complex and dynamic traffic environments. One of the most challenging scenarios for AVs is a cut-in, where another vehicle enters abruptly the ego vehicle's lane, reducing the available time and space for evasive action. Traditional rule-based or reactive collision avoidance systems often struggle in such situations due to the difficulty of predicting the precise timing and nature of the cut-in.

This paper proposes an Intelligent rules-based approach (RBA) model is a deep learning with rules-based approach combined with Time-to-Collision (TTC) metrics to predict potential collisions during cut-in events and to recommend evasive actions. TTC is widely used in collision avoidance systems [19], measuring the time remaining before two vehicles collide based on their relative velocity and distance. However, traditional TTC-based systems may be limited by their static nature, often failing to account for the complex interactions between vehicles in a dynamic traffic environment. By integrating RBA models, our system can learn from diverse traffic situations and predict collisions more effectively, resulting in more timely and appropriate evasive actions.

In cut-in scenarios, a vehicle from an adjacent lane makes a sudden move into the ego vehicle's path, sharply increasing the potential for a collision. The Time-to-Collision (TTC) metric is crucial in these situations, as it estimates the time remaining before an impact by factoring in the relative distance and velocity between the ego vehicle and the cutting-in vehicle. Accurate TTC calculations help the system assess the urgency of the situation and determine whether evasive maneuvers or braking are necessary to prevent an accident.

This paper presents two key contributions: the integration of RBA model with Time-to-Collision (TTC) metrics to improve the accuracy of collision prediction in cut-in scenarios, and the development of a comprehensive collision avoidance strategy. The proposed system not only suggests deceleration but also incorporates lane changes as part of its response, offering a more adaptive and effective solution for preventing accidents in complex traffic conditions.

The goals of this research are to develop a reliable prediction system for detecting potential collisions during cut-in scenarios, propose suitable evasive maneuvers such as deceleration or lane changes, and ultimately enhance the safety of autonomous vehicles (AVs) by improving their collision avoidance capabilities in complex traffic environments.

The novelty of this research lies in combining advanced machine learning techniques with behavioral insights and dynamic TTC computation to create a comprehensive, adaptive collision avoidance framework that addresses limitations in static, rule-based systems. It provides new capabilities for handling unpredictable cut-in scenarios by integrating predictive modeling, human behavior analysis, and multidimensional evasive actions.

This paper is organized as follows: Section 2 provides an overview of related research. Section 3 describes the methodology. Section 4 presents the repository and scenarios used in this study. Sections 5 and 6 discuss the results, conclude the paper, and outline directions for future research.

## 2.      Related Work

Collision avoidance systems for autonomous vehicles (AVs) have evolved from rule-based systems to more sophisticated, machine learning-driven models to handle complex driving scenarios. Rule-based systems follow strict, predefined rules for vehicle control, such as stopping when an object is detected within a specific distance. While simple, these systems struggle in dynamic, real-world scenarios like cut-ins, where the behavior of other vehicles is unpredictable and rapid decisions are required. Machine learning models, such as support vector machines (SVMs), decision trees, and random forests, learn from historical traffic data to predict collision risks by identifying patterns in vehicle behavior [1]. However, these approaches also face challenges in generalizing to real-time, dynamic situations, particularly with cut-ins that require split-second responses.

Yan et. al. [2] focuses on developing driver trust in assistance systems by adapting the system's support to the driver's uncertainty. The premise is that appropriate trust can be fostered when these systems help reduce uncertainty, such as in lane change maneuvers, by adjusting to the driver's uncertainty about distance gaps and closing velocity. The paper presents the creation of a probabilistic model to classify driver uncertainty during lane changes, using data from a simulator experiment. Three Bayesian networks are explored: a naive Bayesian classifier, a Tree-Augmented-Naive Bayesian classifier, and a fully connected Bayesian network [11].   At the same time, some researchers have developed various methods to improve the effect of identifying vehicle lane-changing behavior. Zou et al. [6] proposed a machine learning-based vehicle acceleration prediction model that incorporates driving behavior analysis. By preprocessing driving data and selecting key features like relative distance, velocity, and acceleration, the model aims to improve the accuracy of Advanced Driving Assistance Systems (ADAS) and enhance traffic safety.

Du et al. (2022) [7] developed an intelligent approach to predict lane-change behavior in autonomous vehicles (AVs), using both driving style and trajectory data of AVs and surrounding vehicles. A modified dataset based on real vehicle trajectories (NGSIM) was created for this purpose. The method employs a hidden Markov model (HMM) to assess whether the environment is suitable for lane changes and a learning-based model to predict AV lane changes based on driving conditions. This approach improves the safety and accuracy of AV lane-change maneuvers. Advance in deep learning (DL) have significantly enhanced the ability of AVs to handle such complex scenarios [20]. Long Short-Term Memory (LSTM) networks, a type of recurrent neural network (RNN), have shown great promise in capturing time-series data and temporal dependencies. This is crucial in predicting and reacting to traffic situations, such as cut-ins, where the system needs to understand the evolving relationship between the ego vehicle and surrounding vehicles over time. By learning the sequential nature of traffic events, DL models can more accurately predict potential collisions and determine whether to apply braking or execute a lane change [12].

In addition to deep learning techniques [8], uncertainty analysis plays a key role in improving AV collision avoidance systems. AVs must account for multiple sources of uncertainty, such as the unpredictability of human drivers, noisy sensor data, and environmental factors like weather or road conditions [11]. Techniques like Monte Carlo simulations [5] allow for the assessment of how random variations in these inputs affect collision risk. Bayesian Networks and Hidden Markov Models are also employed to manage uncertainty [18], as they provide probabilistic assessments of future states based on current observations. For example, these models can predict the likelihood of a nearby vehicle performing a sudden lane change or braking unexpectedly, allowing the AV to react appropriately. Li et al. (2018) [9] introduced a feature selection method to predict driver lane change (LC) behavior using naturalistic driving data. The goal was to pinpoint and select the most influential features across different LC scenarios. By applying feature selection, the method reduces the dimensionality of training datasets, eliminating redundant data and improving model efficiency in predicting LC behavior.

Zhao et al. (2022) [4] calculated key vehicle metrics, such as velocity, acceleration, and position, using data like vehicle ID and velocity along the X and Y axes. For clustering driving styles, they derived features including distance headway (DHW), time headway (THW), time to collision (TTC), and the inverse of TTC (ITTC). The study presented an RBA technique that utilizes TTC. This approach boosts efficiency and aids in reducing traffic accidents. Human behavior is one of the largest sources of uncertainty in AV collision avoidance. Human drivers exhibit a wide range of behaviors, from cautious to aggressive, and these variations make it difficult for AVs to predict their actions accurately. Factors such as distraction, fatigue, or emotional state further contribute to unpredictable driving patterns. Autonomous systems must be designed to account for inconsistent human responses, such as delayed reactions to sudden traffic changes or abrupt, unsignaled lane changes during cut-in maneuvers [3]. Machine learning models for human behavior prediction, including advanced machine learning models, are trained on large datasets of human driving patterns. These models help AVs anticipate driver actions, such as lane changes or decelerations, in real-time scenarios. Defensive driving strategies, where AVs proactively maintain safe distances and avoid risky maneuvers, complement these models. By continuously processing real-time data from sensors like LiDAR, cameras, and radar, AVs can adjust their driving strategies to mitigate the risks posed by uncertain human behaviors.

This study focuses on the cut-in scenario, where another vehicle unexpectedly merges into the ego vehicle's lane, requiring a swift response to avoid a collision. Critical parameters such as reaction time, maximum deceleration, and velocity are essential in evaluating the ego vehicle's ability to respond safely and effectively to these sudden changes in traffic dynamics. In this context, Time-to-Collision (TTC) is a vital metric that estimates the remaining time until a collision would occur if both vehicles continued at their current velocity and paths. TTC metric [10], a well-established predictor, calculates the remaining time before a collision based on the current velocity and distance between vehicles. While useful for basic predictions, traditional TTC-based methods falter in rapidly changing situations, such as sudden lane changes or unpredictable cut-ins, where a collision can occur if evasive actions are not timed perfectly. TTC assumes constant vehicle velocity, making it less effective when the relative velocity between vehicles fluctuates or when lane positions change abruptly.

## 3.    Methodology

### 3.1.    Data Collection

The system is developed and validated using a simulated dataset that captures a wide variety of cut-in scenarios. The data are from the highD datasets [21]. The human driver trajectories in the data base are collected on German highways at six different locations near cologne using unmanned aerial vehicles. The dataset includes crucial information such as vehicle dynamics, covering the positions, velocities, and accelerations of both the ego vehicle and the cutting-in vehicle. Additionally, it incorporates details about traffic conditions, including road types, traffic density, and overall traffic flow. To ensure the model's robustness, various types of cut-in events are represented, ranging from abrupt and gradual to emergency maneuvers. Figure 1 describes the algorithm of RBA avoidance for cut-in safety.

**Input:**

Ego vehicle: position, velocity, width, length, deceleration, reaction time, safety distance.

Cutting-in vehicle: position, velocity, width, length.

**Step 1**: Lateral Safety

Calculate lateral distance and subtract half the width of both vehicles.

Compare to ego vehicle's safe lateral distance to eventually presenter or (from behind) approaching car on LC-target lane. If greater, proceed; otherwise, unsafe.

**Step 2**: Longitudinal Safety

Calculate longitudinal distance, subtracting half the lengths of both vehicles.

Adjust safe distance dynamically based on ego vehicle velocity, provided the sensors have sufficient predictive horizon backwards, i.e. can reliably recognize the approaching vehicle(s) from behind. If greater, proceed; otherwise, unsafe.

**Step 3**: Velocity Difference

Calculate velocity difference for stopping time margin.

**Step 4**: Stopping Time Margin

Compute stopping time margin based on velocity, deceleration, and reaction time.

If the ratio of longitudinal distance to velocity difference exceeds the margin, proceed.

**Step 5**: Safety Check

If lateral, longitudinal, and stopping time checks pass, cut-in is safe; otherwise, unsafe.

**Output:**

Return True for safe, False for unsafe.

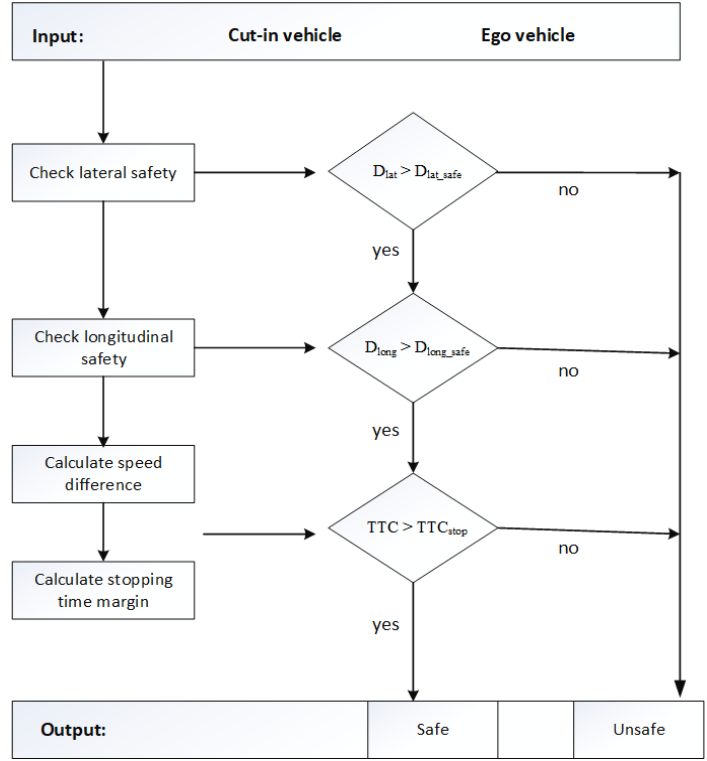

**Figure 1**.　Algorithm description, provided sensor coverage (for Level 3 AV) is available

*3.2 Mathematical description of the Collision Avoidance Logic*

The safety conditions (eq.1 and velocity adjustment eq.2 can be expressed as follows:

$$Safe = \left( \left| L_{ego,lat}(i) - L_{cut,lat}(i) \right| - \frac{W_{ego} + W_{cut}}{2} > D_{lat,safe} \right) \wedge \left( \left| L_{ego}(i) - L_{cut}(i) \right| - \frac{L_{ego\_veh + L_{cut\_veh}}}{2} > D_{safe} \vee \frac{\left| L_{ego}(i) - L_{cut}(i) \right| - \frac{L_{ego\_veh + L_{cut\_veh}}}{2}}{\left| V_{ego}(i) - V_{cut}(i) \right|} > \frac{V_{ego}(i) - V_{cut}(i)}{2 \times A_{max}} + T_{react} + 0.1 \right) \quad (1)$$

If Safe is false, adjust the velocity:

$$\Delta V_{dec} = \min \left( \frac{\max \left( \frac{D_{safe} + \text{safety buffer} - D_{long}}{D_{safe} + \text{safety buffer}}, \frac{TTC_{safe} - TTC}{TTC_{safe}} \right) \times A_{max}}{f}, A_{max} \right) \quad (2)$$

$$V_{ego}(i+1) = \max \left( V_{ego}(i) - \Delta V_{dec}, V_{min} \right) \quad (3)$$

**Table 1**. The following abbreviations are used in this manuscript

| $L_{ego}(i)$ | Longitudinal position of the ego vehicle at time step i |
|---|---|
| $L_{cut}(i)$ | Longitudinal position of the cutting-in vehicle at time step i |
| $L_{ego,\ lat}(i)$ | Lateral position of the ego vehicle at time step i |
| $L_{cut,\ lat}(i)$ | Lateral position of the cutting-in vehicle at time step i |
| $V_{ego}(i)$ | Longitudinal velocity of the ego vehicle at time step i |
| $V_{cut}(i)$ | Longitudinal velocity of the cutting-in vehicle at time step i |
| $W_{ego}$ | Width of the ego vehicle |
| $W_{cut}$ | Width of the cutting-in vehicle |
| $L_{ego\_veh}$ | Length of the ego vehicle |
| $L_{cut\_veh}$ | Length of the cutting-in vehicle |
| $W_{ego}$ | Width of the ego vehicle |
| $W_{cut}$ | Width of the cutting-in vehicle |
| $L_{ego\_veh}$ | Length of the ego vehicle |
| $L_{cut\_veh}$ | Length of the cutting-in vehicle |
| $W_{ego}$ | Width of the ego vehicle |
| $W_{cut}$ | Width of the cutting-in vehicle |
| $L_{ego\_veh}$ | Length of the ego vehicle |
| $L_{cut\_veh}$ | Length of the cutting-in vehicle |
| $W_{ego}$ | Width of the ego vehicle |

### 3.3 Feature extraction

It is first necessary to obtain the features that can represent the driving style. In this paper, the main extracted features include the distance headway (DHW), time headway (THW), time to collision (TTC), and the inverse of TTC (ITTC). The DHW represents the distance between the front and rear vehicles. The THW represents the time difference between the front and rear vehicles passing through the same place; it can be calculated by dividing the DHW by the following vehicle velocity. The TTC indicates the time required for the collision if two vehicles continue to collide at the current velocity and on the same path; it can be calculated by dividing the DHW by the velocity difference between two vehicles.

• Cut-In Scenario: In a cut-in scenario, the collision avoidance system evaluates four key parameters simultaneously to determine the most effective response:

• Ego Vehicle Velocity ($V_{e0}$): Influences the reaction time available to the system to handle a cut-in situation.

• Lateral Distance ($d_{y0}$): Identifies the degree of lane intrusion by the cutting-in vehicle, helping detect when it enters the ego vehicle's path.

• Longitudinal Distance ($d_{x0}$): Essential for calculating the Time-to-Collision (TTC), assessing if the ego vehicle has sufficient time to avoid an impact.

• Lateral Velocity ($V_y$): Indicates how quickly the cut-in is happening, guiding the system on how urgently it needs to act.

By continuously monitoring these parameters in real-time, the system predicts collision risks based on threshold as illustrates Figure 2 and takes proactive measures, ensuring safe navigation through complex traffic scenarios.

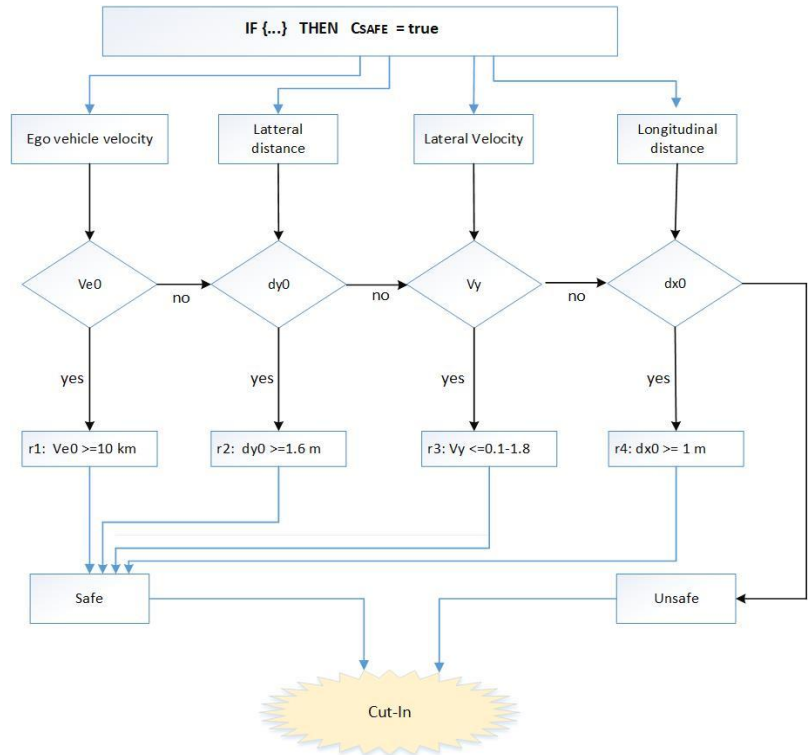

**Figure 2.** Cut-In scenario

### 3.4 Safety-Critical Driving Situations in Cut-In Scenarios

Safety-critical situations arise when the risk of collision becomes imminent, demanding immediate action to prevent an accident. These scenarios are marked by the rapid reduction of available space and time for the ego vehicle to respond. Key characteristics of such situations include sudden or aggressive maneuvers by a cutting-in vehicle, which present a direct threat to the safety of the ego vehicle and other road users. Specific examples include a vehicle making a sudden cut-in at high velocity, reducing the gap to the ego vehicle and necessitating quick deceleration or lane changes. Another example is when a vehicle cuts in and then brakes abruptly, leaving little time for the ego vehicle to react safely. In poor weather conditions, such as rain, snow, or fog, a cut-in becomes even more dangerous, as reduced visibility and traction increase the stopping distance. In these scenarios, advanced collision avoidance systems leveraging time-to-collision (TTC) metrics combined with deep learning models must swiftly predict the collision risk and execute appropriate evasive maneuvers.

### 3.6. Non-Safety-Critical Driving Situations in Cut-In Scenarios

Non-safety-critical situations occur when a vehicle cuts into the ego vehicle's lane in a gradual manner, allowing ample time and space for the ego vehicle to react without an immediate threat of collision. In these cases, while adjustments may still be required, the urgency is low, and evasive maneuvers are often unnecessary. For example, in a gradual cut-in with sufficient space, the cutting-in vehicle leaves enough room for the ego vehicle to maintain a safe following distance with a slight reduction in velocity. Similarly, in low-velocity traffic, cut-ins pose minimal risk as both vehicles have enough time to adjust. A cut-in on an empty road provides even more leeway, as the ego vehicle can easily change velocity or lane position. Lastly, when a cut-in is anticipated or predicted, such as through early lane-change signals, the ego vehicle can smoothly adapt without a sense of urgency. Table 2 compares the critical and nob critical safety events in cut-in scenario.

Table 2:Critical vs. Non-Critical Safety Events

| Component | Critical Safety Events | Non-Critical Safety Events |
|---|---|---|

| Perception | Immediate threat detection, rapid object tracking, immediate response required | Routine object detection, tracking with no urgency |
|---|---|---|
| Decision | Split-second risk assessment, emergency planning | Long-term planning, efficiency-driven decisions |
| Reaction | Urgent, precise control (e.g., hard braking), lane change | Smooth, controlled actions (e.g., gradual slowing), sufficient space and time to react |
| Focus | Avoiding or mitigating collisions | Maintaining safe, comfortable driving behavior |
| Driver/System Action | Aggressive evasive maneuvers required | Minor velocity adjustments or no action needed |
| Driving Conditions | Often in dense traffic, high velocitys, or poor weather | Usually in open or slow-moving traffic conditions |
| Time Sensitivity | Real-time or near-real-time | Longer decision window |

### 3.5 Collision Avoidance Strategy

The trained deep learning model predicts time-to-collision (TTC) values to assess whether a collision is imminent. Upon detecting a potential collision, the system determines the most effective evasive action, which may include deceleration or a lane change. If the predicted TTC indicates an imminent threat, the system calculates the required deceleration to reduce the velocity of the ego vehicle to avert the collision. If deceleration alone is insufficient or impractical due to surrounding traffic conditions, the model suggests a safe lane change. The collision avoidance system carefully evaluates both options, considering current traffic dynamics and vehicle capabilities, to ensure that the chosen action effectively mitigates the risk of a collision while maintaining overall safety on the road.

Simulations are conducted using various cut-in scenarios, including highway driving with sudden cut-ins, urban traffic with frequent lane changes, and emergency braking situations, to evaluate the system's performance. The effectiveness of the deep learning-based system is compared to traditional TTC-based methods, highlighting the advanced model's improved adaptability and accuracy. Uncertainty analysis assesses how variations in input parameters, such as measurement errors, environmental conditions, and human driving behavior, impact the system's ability to predict and prevent collisions. By identifying these uncertainties, engineers can enhance system robustness and ensure more reliable risk assessment. Sensitivity analysis evaluates how changes in key parameters, like vehicle velocity or reaction time, influence the system's output, helping to identify which factors are most critical as illustrates Figure 3. Combining uncertainty and sensitivity analysis provides a comprehensive understanding of the model's behavior, allowing improvements in system design, sensor fusion, and real-time adaptability to maximize safety in autonomous driving systems.

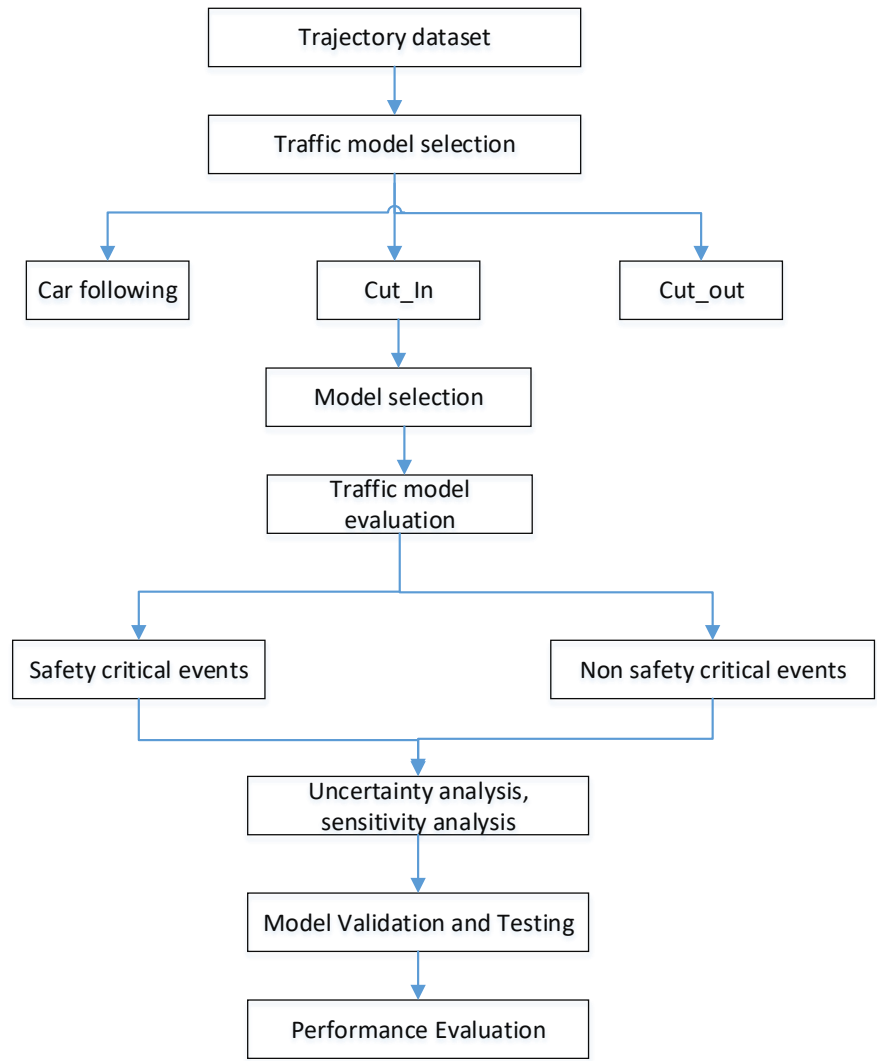

**Figure 3.** Sensitivity analysis

*3.6 Sensitivity Analysis in Collision Avoidance Systems*

In cut-in scenarios as illustrates Figure 4, sensitivity analysis is crucial for determining how different input parameters impact the behavior of a collision avoidance system. Key parameters typically include Time-to-Collision (TTC), which measures the time remaining before a collision if vehicles maintain their current velocity and trajectory, and Relative Velocity, the difference in velocity between the ego and cutting-in vehicles, influencing collision severity. Inter-vehicle Distance and cut-in angle are also critical, as shorter gaps and sharper angles increase collision risk. Reaction time is tested to see how delays in detecting and responding to cut-ins affect outcomes, while weather and road conditions, such as rain or slippery surfaces, further complicate the system's response. Driver behavior uncertainty, like inattentiveness or fatigue, is another factor that can affect reaction velocity. By analyzing these parameters, developers can enhance the robustness of collision avoidance systems, ensuring more reliable performance in real-world situations.

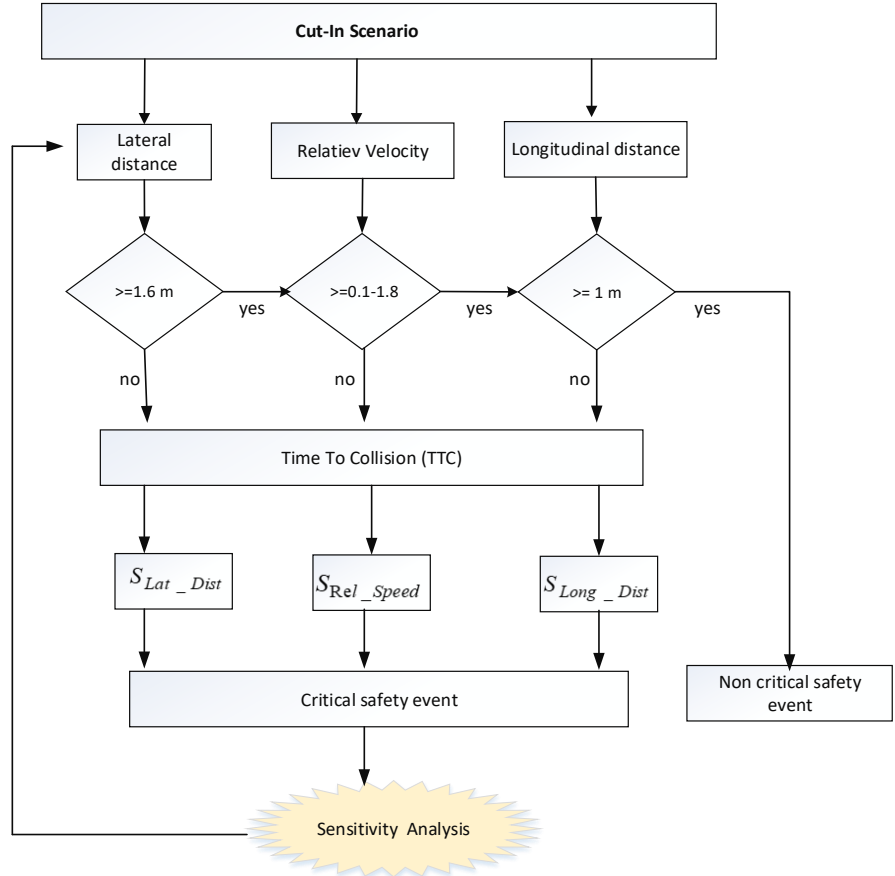

**Figure 4.** Cut-In scenario

### 3.7 . Sensitive of safety metrics

Sensitivity analysis helps evaluate how the variation in the output of a model can be attributed to different variations in its inputs. It is especially useful in complex systems where understanding the relationships between inputs and outputs can improve decision-making, optimization, and robustness. When a vehicle performs a cut-in maneuver, changes in key parameters such as lateral distance, relative velocity, and longitudinal distance can significantly affect safety metrics like Time-to-Collision (TTC). Lateral distance, relative velocity, and longitudinal distance all play crucial roles in determining the safety and time-to-collision (TTC) during a cut-in maneuver. A smaller lateral distance indicates a more aggressive lane change, reducing reaction time and increasing the risk of collision, while a larger lateral gap allows for smoother adjustments, lowering the immediate danger. Relative velocity is another key factor; if the cut-in vehicle is slower, the closing rate rises, requiring quicker reactions from the host vehicle to avoid a crash, whereas a faster cut-in vehicle reduces risk by moving away. A higher relative velocity shortens TTC, increasing collision likelihood unless the host decelerates, while a negative relative velocity extends TTC. Finally, longitudinal distance directly affects safety, with a smaller gap leaving less time for reaction and reducing TTC, whereas a larger gap provides more time to react and maintain safety. To measure the sensitivity of safety metrics, such as Time-to-Collision (TTC), to changes in inputs like longitudinal distance, relative velocity, and lateral distance, a systematic approach is required. Which means, sensitivity can be measured by how much TTC changes in response to a change in the input parameter such as Time-to-Collision (TTC). This can be done using partial derivatives (for continuous sensitivity) or by percent change (for discrete changes). To calculate instantaneous sensitivity, use partial derivatives of TTC with respect to each input.

- *Sensitivity to longitudinal distance*

$$S_{long\_Dist} = \frac{\partial TTC}{\partial d} \qquad (4)$$

• Sensitivity to relative velocity

$$S_{Rel\_Speed} = \frac{\partial TTC}{\partial v_{rel}} \qquad\qquad (5)$$

*3.8   Lateral Distance Sensitivity*

While lateral distance does not directly impact time-to-collision (TTC), it indirectly influences the host vehicle's reaction time. A larger lateral distance can provide the host vehicle with additional time to react, effectively increasing the TTC, while a smaller lateral distance demands quicker responses, leading to a reduced TTC as illustrated in Figure 5. In contrast, longitudinal distance and relative velocity directly affect TTC longitudinal distance is proportional to TTC, whereas relative velocity is inversely related. Lateral distance indirectly shapes TTC by determining how swiftly the host vehicle can respond to a cut-in, with sensitivity increasing at lower longitudinal distances and higher relative velocities, particularly when relative velocity is low.

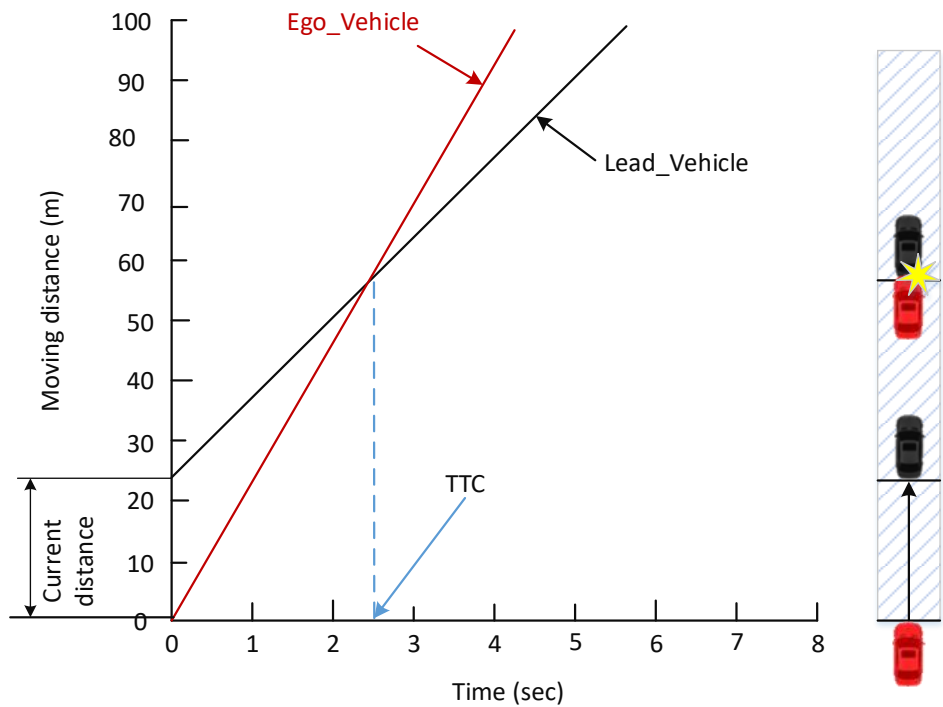

Figure 5: Variation of Time-to-Collision (TTC)

## 4.   Safety Driving Simulation

In this section, we describe the repository used in the current work [22].
The implementation includes three reference scenarios:

• cut_in
• cut_out
• car_following

Four safety models are supported:

• FSM – Fuzzy Safety Model
• RSS – Responsibility Sensitive Safety
• CC_human_driver
• Reg157

The Python script safety_check_runner.py enables three types of analyses:

- one_case – allows selecting a specific scenario, model and provides visual inspection of the simulation results.
- comparison – performs a systematic evaluation of the safety models over a set of logical scenarios and stores the results for further processing.
- post_processing – enables visual inspection of the results generated during a previous comparison run.

The script safety_check_runner.py can be executed either from the command line, using the minimum required arguments, or programmatically from another Python script, as demonstrated in example.py.

- Cut-In scenario

In this study, we focus on the cut-in scenario. Table 3 summarizes the key parameters required to describe the pattern of traffic-critical situations in this scenario. Additional parameters may be incorporated depending on the operating environment (e.g., road friction, road curvature, lighting conditions). Figure 6 provides a visual illustration of the cut-in scenario.

Table 3. parameter for cut-in scenario

| parameters | description | Units |
|---|---|---|
| $V_{e0}$ | Ego vehicle | m/s |
| $V_{o0}$ | Leading vehicle in lane or in adjacent | m/s |
| $d_{x0}$ | Longitudinal distance | m |
| $d_{y0}$ | lateral distance | m |

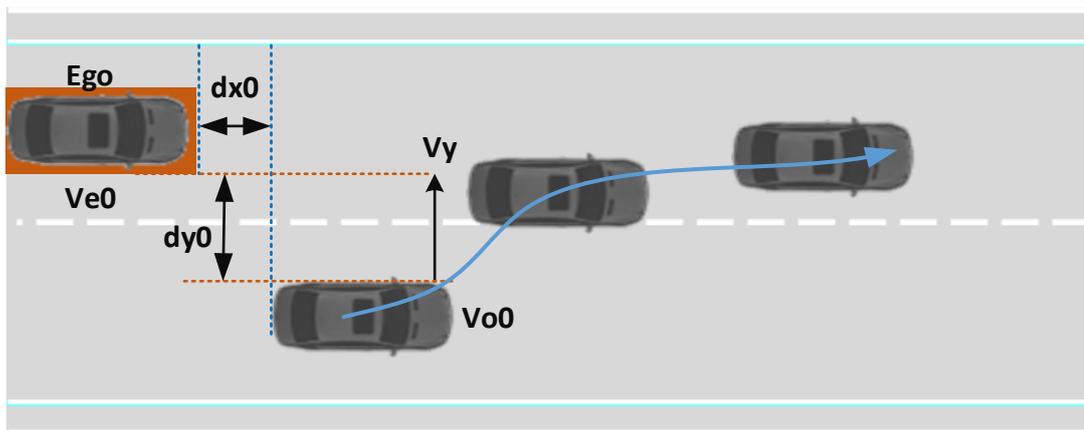

Figure 6: Cut-in scenario

- Algorithm Pseudocode and Vehicle Data

Figure 7 presents the safety and reaction-time functions implemented in the repository, along with the corresponding vehicle data.

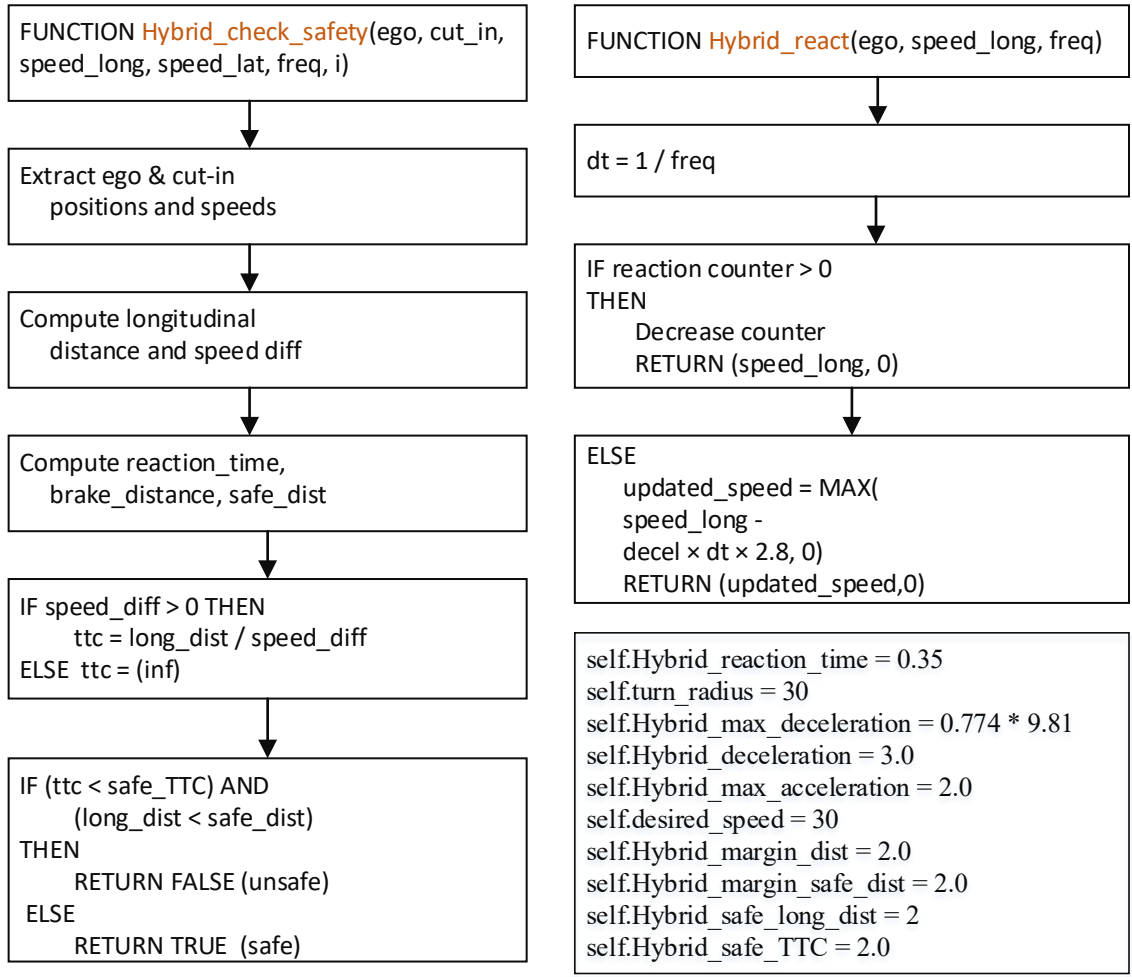

**Figure 7:** Safety scheme and vehicle data

## 5. Results and Discussion

The simulation results demonstrate that the RBA-based collision avoidance system outperforms traditional TTC-based methods. The system accurately predicts potential collisions and initiates timely evasive actions, significantly reducing the risk of collisions in cut-in scenarios. The simulation results demonstrate that the RBA-based collision avoidance system outperforms traditional TTC-based methods. The system accurately predicts potential collisions and initiates timely evasive actions, significantly reducing the risk of collisions in cut-in scenarios.

Time-to-Collision (TTC) is a critical measure in assessing collision risk, as it represents the time remaining before a collision occurs if the velocity of both vehicles remains constant. When a cut-in vehicle moves significantly slower than the ego vehicle, the TTC decreases, meaning the distance between them closes faster, increasing the collision risk as illustrates Figure 8. However, as the cut-in vehicle's velocity approaches that of the ego vehicle, the TTC increases, providing more time for the ego vehicle to react and adjust velocity or position, thus reducing the chance of a collision. Additionally, with a smaller relative velocity, the required longitudinal safe distance decreases, allowing the ego vehicle to maintain a safe following distance without harsh braking. This smoother driving dynamic also improves reaction time, enabling the ego vehicle to make gradual adjustments, ultimately reducing the risk of rear-end collisions and enhancing overall road safety.

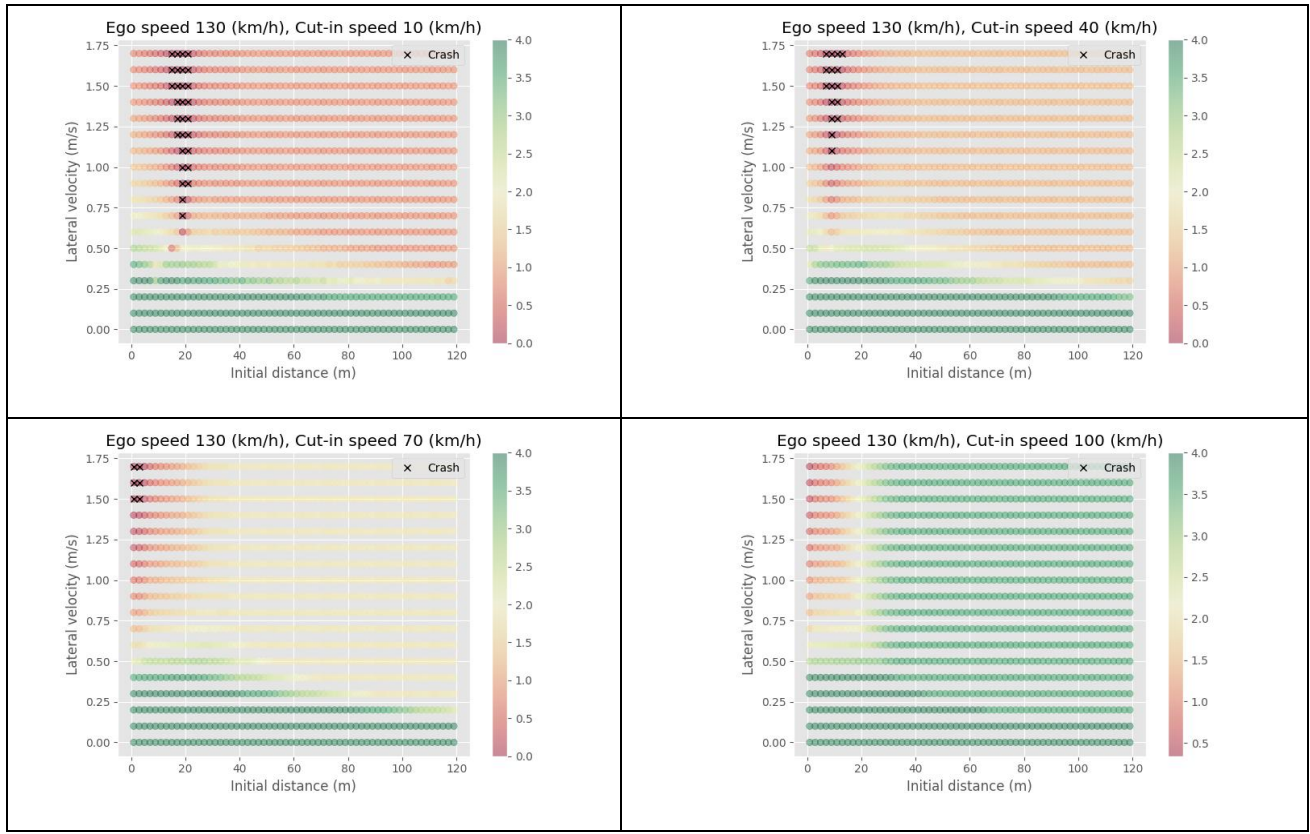

**Figure 8**. Ego velocity vs. Cut-In object velocity

*5.1 Comparative Analysis of Collision Avoidance Models*

To analyze and compare various collision avoidance methods—RBA, Responsibility-Sensitive Safety (RSS), Regulation 157 (Automated Lane Keeping Systems), FSM, IDM, CC human driver based on ego velocity and cut-in velocity as illustrates Figure 9, we can evaluate how each method handles velocity dynamics, safe distances, and reactions to cut-in events. Here we follow the procedures formulated in [13]. We use the github-repository to investigate the behavior of the proposed safety models for UNECE Reg157. The repository discusses four models: Fuzzy Safety Model (FSM) [14], Responsibility Sensitive Safety (RSS)[15], CC human driver [16], and Reg 157 [17]. In addition to these repository models, we include our RBA model, which has been described in section 3. Methodology, RBA Collision Avoidance for Cut-In Safety. We compare the results as shown in Figure 9. The repository implements three reference scenarios. We focus on cut-in, which carries the biggest risk of collision. The python script 'safety_check_runner.py' provides the possibility of selecting of logical scenario; and 'post_processing' to visually inspect the results of the previously executed 'comparison' scenario. The key difference between these approaches lies in how they interpret Time-to-Collision (TTC), safe distances, and braking strategies during cut-in scenarios. The RBA model dynamically adjusts lateral and longitudinal safety thresholds, prioritizing smooth braking and minimal disruption when the cut-in velocity closely matches the ego vehicle. RSS, on the other hand, formalizes safe distances based on legal frameworks and ensures minimal deceleration when velocitys are similar, while increasing buffer zones at lower cut-in velocitys. Regulation 157, designed for lower-velocity environments, focuses on maintaining a safe following distance in urban traffic but is limited in high-velocity scenarios. CC human driver adopts a highly cautious approach, often leading to early braking and overly conservative behavior even when the velocity difference is small, resulting in less optimal comfort and driving smoothness compared to RBA or RSS.

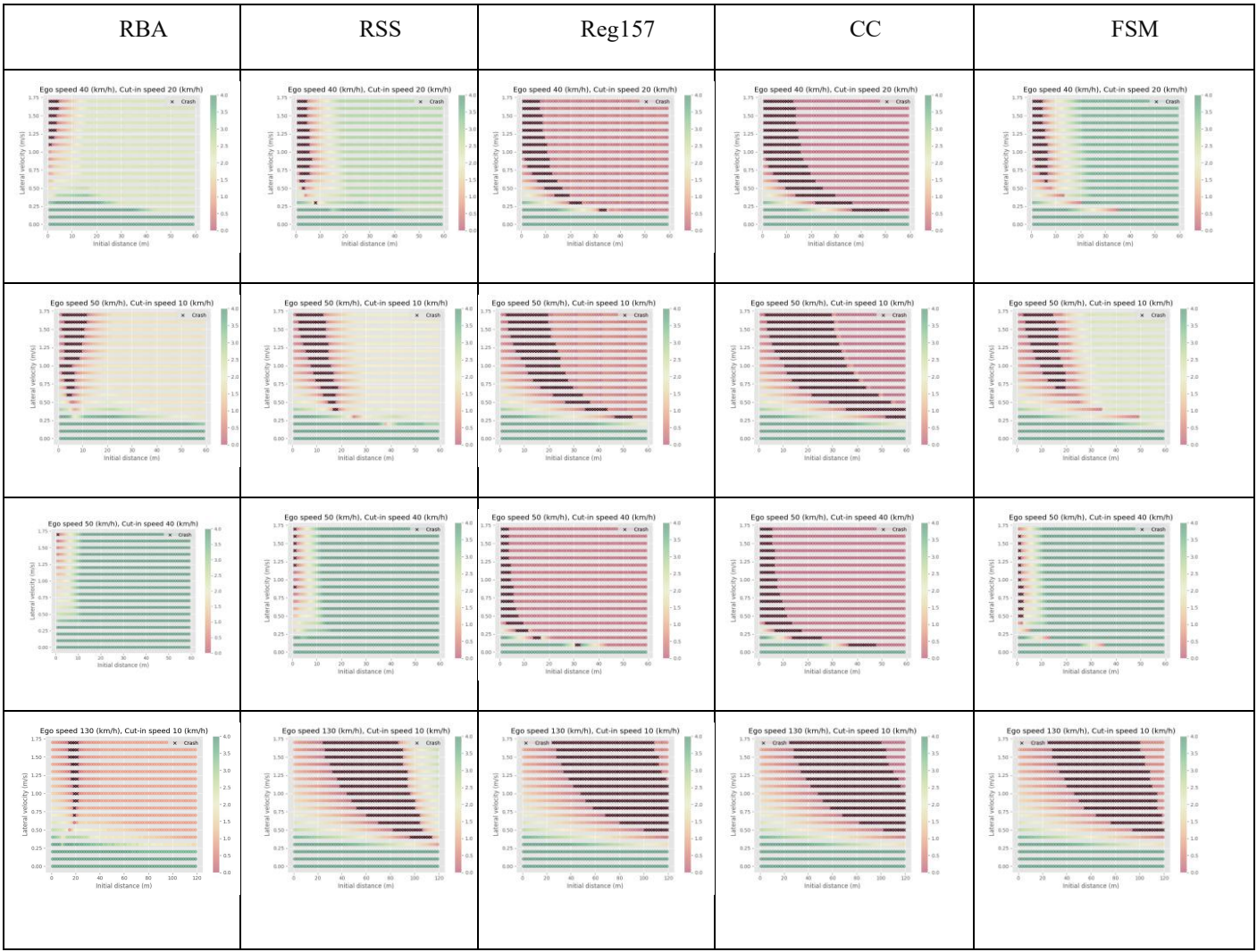

**Figure 9.** Analysis of Collision Avoidance

*5.2 Ego velocity analysis in Urban*

To analyze Time to Collision (TTC) values based on a Gaussian distribution in the context of collision risk, the mean (μ) represents the average TTC. A high mean (e.g., >5 seconds) suggests the vehicle maintains a safer distance from potential collisions, while a low mean (e.g., <2 seconds) indicates more frequent operation in risky conditions. The standard deviation (σ) measures the spread of TTC values; a low σ means the values are closely clustered, indicating consistent intervals, whereas a high σ shows greater variability, with the vehicle occasionally getting closer to collisions as illustrates in Figure 10. The left tail of the Gaussian curve represents dangerous TTC values near zero, indicating higher collision risk, while the right tail shows safer situations. Analyzing the area under the curve left of a critical threshold (e.g., 2 seconds) gives the proportion of time the vehicle is at high risk. A narrow, tall Gaussian curve suggests stable risk levels, while a wide, flat curve indicates variability between safe and risky TTC values. Comparing the *mean* and *standard deviation* of different systems (e.g., CC, FSM, IDM, Reg157, RSS, Hybrid) helps identify which systems maintain safer TTC (high μ, low σ) and which exhibit higher collision risks (low μ, high σ) as Illustrated in Figure 11. The updated TTC statistics reveal a pronounced separation between models with low mean TTC values and those with consistently high estimates. Interpreting TTC as an indicator of error, risk, or deviation from a safety margin, the models CC (Mean = 0.37) and Reg157 (Mean = 0.50) remain closest to the safety threshold, suggesting more conservative or stable behavior during cut-in events. In contrast, IDM (Mean = 3.90), FSM (Mean = 3.65), RSS (Mean = 3.67), and RBA (Mean = 3.76) all show substantially higher mean TTC values, indicating stronger deviations that may reflect more reactive or more conservative responses depending on the interpretation of the metric. Additionally, the very low variance observed for IDM (Std Dev = 0.40) and RBA (Std Dev = 0.69) highlights their high consistency, whereas models like CC and Reg157 show more variability.

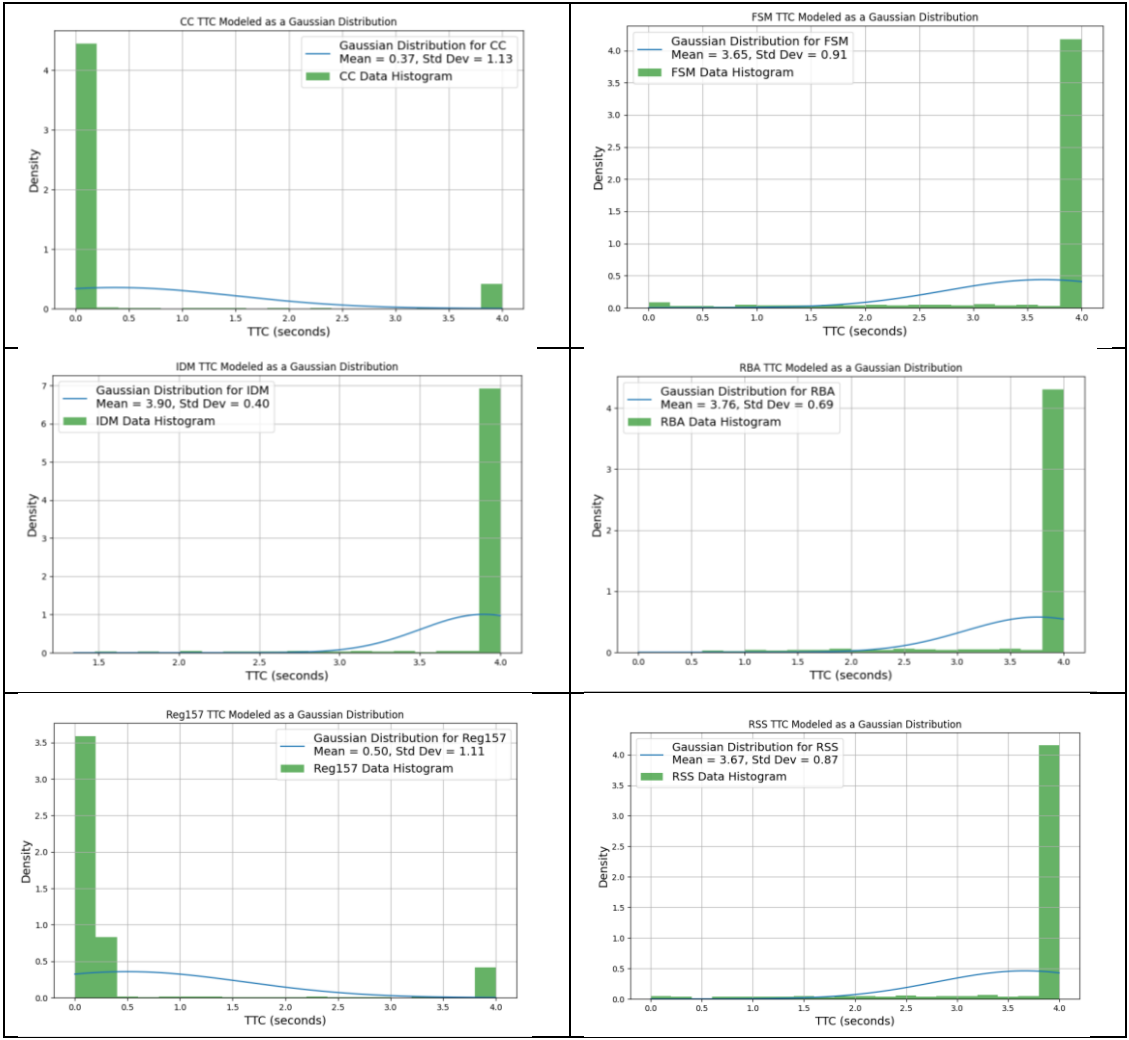

**Figure 10**: TTC modeled as Gaussian Distribution in urban

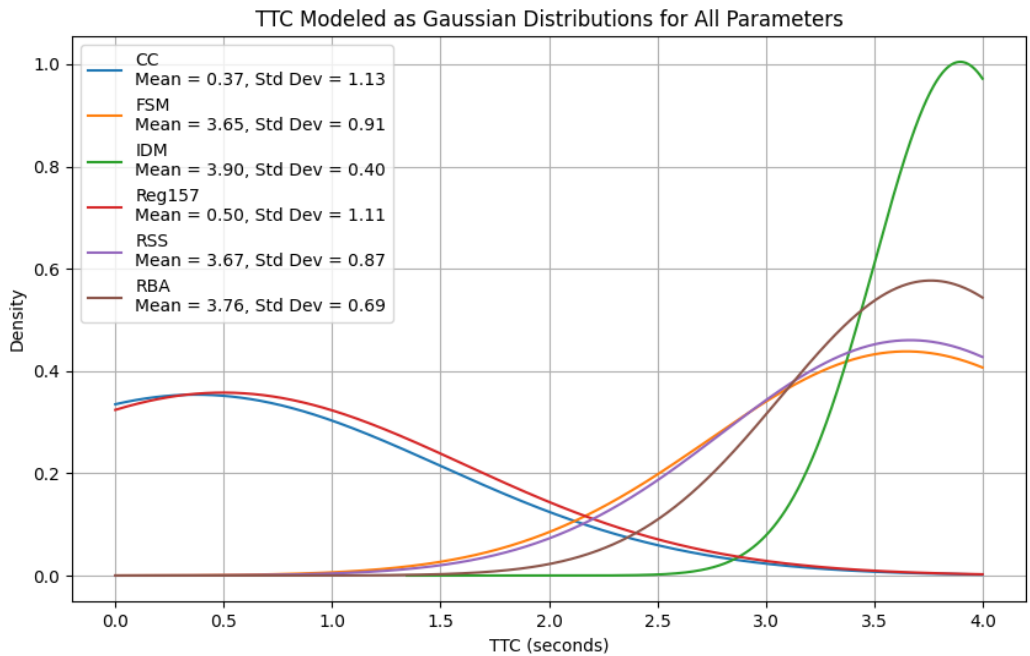

**Figure 11**: TTC modeled as Gaussian Distribution for all parameters

*5.3 Ego velocity analysis in Highway*

Figure 12 illustrates the mean the standard deviation for each model based on Time-to-Collision (TTC) and Gaussian Width of Curves highlights key performance differences. Based on the updated TTC values, a clear distinction emerges between models that exhibit lower average TTC and those that consistently produce higher TTC estimates. If TTC is interpreted as an indicator of error, risk, or deviation from a safety threshold, then the models CC, Reg157, and IDM, which show comparatively low mean TTC values, appear to operate closer to an optimal or safer target region. Their behavior suggests conservative or stable responses that avoid large deviations in critical situations. In contrast, FSM, RSS, and RBA, all showing substantially higher mean TTC values, demonstrate systematically elevated responses. Depending on the interpretation of the metric, these values may reflect either more aggressive, more conservative, or more sensitive model behavior. Alternatively, if the TTC values represent reaction strength, braking demand, or responsiveness to cut-in events, then models with higher means namely RSS, RBA, and FSM can be understood as more reactive to dynamic changes, showing stronger or earlier responses to potential conflicts. Models with lower mean TTC, CC, Reg157, and IDM may instead indicate smoother or less abrupt behavior, responding more passively to the same stimuli as illustrated in Figure 13.. This separation highlights fundamental behavioral differences across the model families and provides insight into their tendencies under critical-driving conditions.

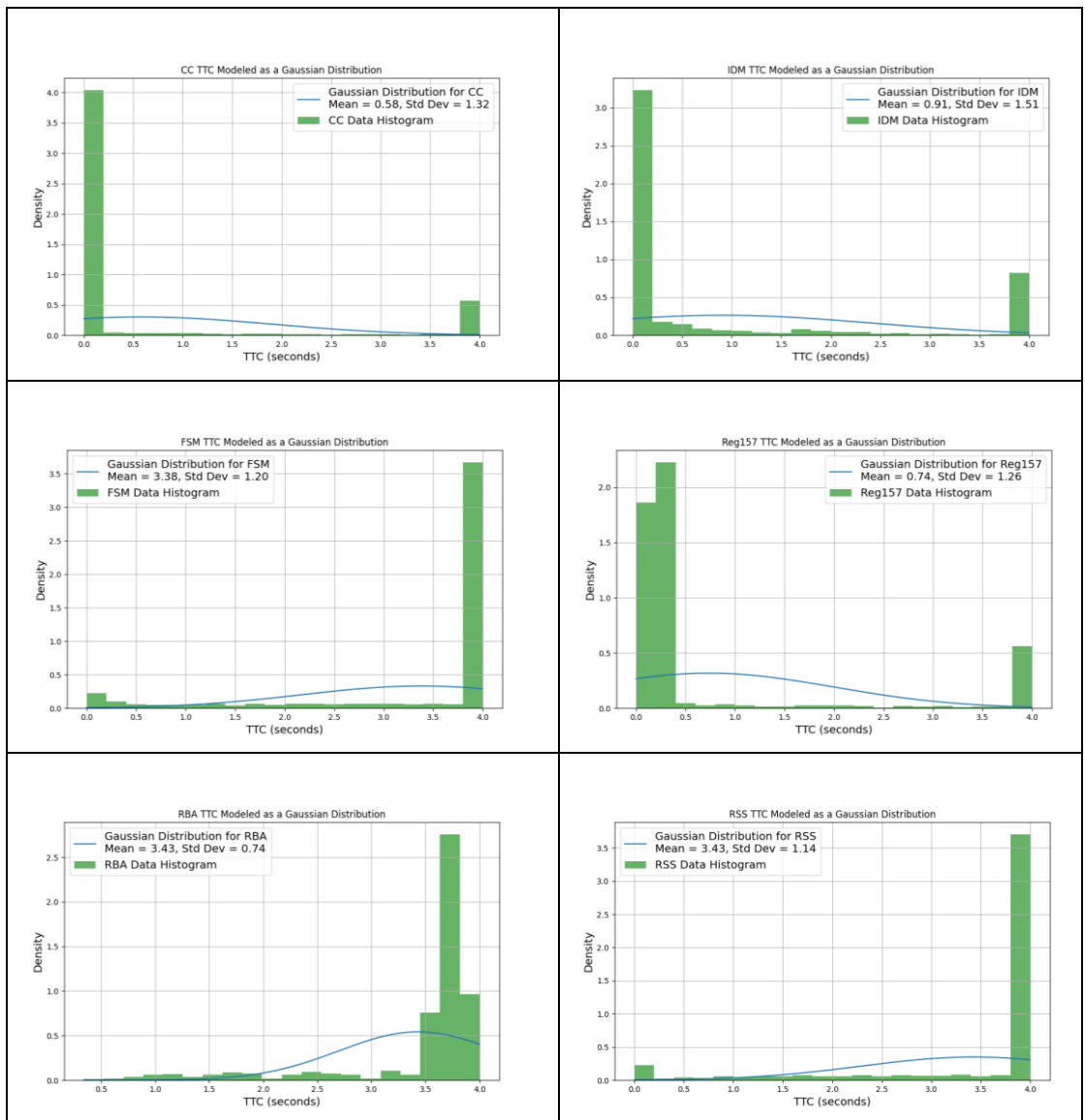

Figure 12: TTC modeled as Gaussian Distribution in Highway

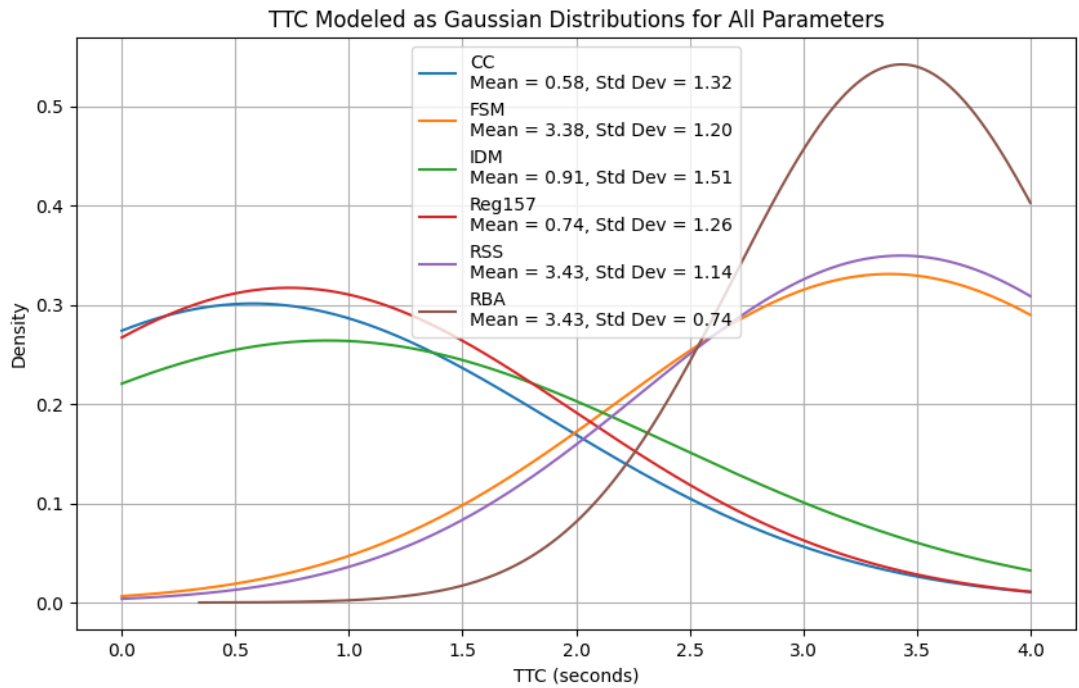

**Figure 13**. TTC modeled as Gaussian distribution for all parameters in Highway

## 6. Conclusion

This paper presents a RBA model-based collision avoidance system for autonomous vehicles, specifically targeting cut-in scenarios. By leveraging TTC and advanced predictive models, the proposed system significantly improves road safety. The methodology and simulation results demonstrate the potential of RBA model in enhancing collision avoidance strategies for autonomous driving. The sensitivity analysis reveals that when relative velocity is low, small changes in velocity have a significant impact on time-to-collision (TTC) because even slight increases in velocity drastically reduce the time to potential impact. Conversely, at higher relative velocity, changes in velocity have a smaller effect on TTC since the sensitivity decreases as the square of relative velocity grows. This insight is crucial for designing driver assistance systems and collision-avoidance algorithms, where understanding how TTC responds to velocity variations is essential for enhancing vehicle safety and preventing collisions. Future work will focus on validating the proposed system in real-world scenarios, integrating additional sensor data to improve prediction accuracy, and exploring advanced multi-agent strategies to address the complexities of dynamic traffic environments, further advancing the safety and adaptability of autonomous driving technologies.

**Acknowledgments:** We are grateful to the Smart Mobility Project at the European Commission, Joint Research Center and in particular to Konstantino Mattas and Biagio Ciuffo for introducing us to the regulations, to the open research questions and for sharing with us the repository of the code, which we used to integrate our model for comparison.

**Declaration of generative AI and AI-assisted technologies in the writing process**
During the preparation of this paper, the author used ChatGPT 4 to conduct the proofreading and correct language errors wherever necessary. After using this tool/service, the author reviewed and edited the content as needed and takes full responsibility for the content of the publication.

**Declaration of Competing Interest**
The authors declare that they have no known competing financial interests or personal relationships that could have appeared to influence the work reported in this paper.

**Conflicts of Interest:** The authors declare no conflict of interest

Data availability
**Data Availability Statement:** Data will be made available on request.

Code availability
The code for the repository is available as Code (https://github.com/ec-jrc/JRC-FSM)